\pdfoutput=1
\documentclass[11pt]{article}

\usepackage[table]{xcolor}

\usepackage{emnlp2021}

\usepackage{times}
\usepackage{latexsym}
\usepackage{todonotes} % for some baffling reason, tables and figures get messed up if this package (which is unrelated to tables and figures) isn't loaded.
\usepackage{soul}

\usepackage{booktabs}
\usepackage{amsmath}
\usepackage{enumitem}
\usepackage{multirow}
\usepackage{multicol}
\usepackage{float}
\usepackage{pdfcomment}
\usepackage{tabularx}

\usepackage{scalerel,amssymb}
\def\chkbox{\mathord{\scalerel*{\Box}{gX}}}

\usepackage{microtype}

\title{Does Putting a Linguist in the Loop Improve NLU Data Collection?}

\author{Alicia Parrish,$^1$ 
        William Huang,$^1$
        Omar Agha,$^1$
        Soo-Hwan Lee,$^1$
        Nikita Nangia,$^1$\\\bf
        Alex Warstadt,$^1$ 
        Karmanya Aggarwal,$^2$
        Emily Allaway,$^3$
        Tal Linzen,$^1$
        Samuel R. Bowman$^1$ \AND
\textnormal{$^1$New York University} \And
\textnormal{$^2$IIIT-Delhi} \And
\textnormal{$^3$Columbia University} \AND 
Correspondence: {\tt \{\href{mailto:alicia.v.parrish@nyu.edu}{alicia.v.parrish},
\href{mailto:nikitanangia@nyu.edu}{nikitanangia},
\href{mailto:linzen@nyu.edu}{linzen}, \href{mailto:bowman@nyu.edu}{bowman}\}@nyu.edu}}

\date{}

\begin{document}
\maketitle
\begin{abstract} % 185 words (of a 200 word limit)

Many crowdsourced NLP datasets contain systematic gaps and biases that are identified only after data collection is complete.
Identifying these issues from early data samples during crowdsourcing should make mitigation more efficient, especially when done iteratively.
We take natural language inference as a test case and ask whether it is beneficial to put a linguist `in the loop' during data collection to dynamically identify and address gaps in the data by introducing novel constraints on the task. 
%These challenges require the crowdworker to write examples that mitigate these issues. 
We directly compare three data collection protocols: (i) a baseline protocol, (ii) a linguist-in-the-loop intervention with iteratively-updated constraints on the task, and (iii) an extension of linguist-in-the-loop that provides direct interaction between linguists and crowdworkers via a chatroom.
%For issues that we explicitly targeted with bonus challenges, 
The datasets collected with linguist involvement are more reliably challenging than baseline, without loss of quality. 
But we see no evidence that using this data in training leads to better out-of-domain model performance,
and the addition of a chat platform % for direct, one-on-one interaction between experts and crowdworkers 
has no measurable effect on the resulting dataset.
%We suggest that dynamic, expert-guided interventions lead to more robust data collection for in-domain metrics, but continued one-on-one interaction between experts and crowdworkers was not helpful.
%We suggest that it is beneficial to integrate expert analysis \textit{during} data collection, so that the expert can dynamically address existing gaps and biases in the dataset.
We suggest integrating expert analysis \textit{during} data collection so that the expert can dynamically address gaps and biases in the dataset.

\end{abstract}

\section{Introduction} % Alicia, Alex, Soo-Hwan, Nikita

\begin{figure}[h!]
    \centering
    %\pdftooltip{\includegraphics[width=0.85\linewidth]{figs/protocols_diagram_version9.1.png}}{A diagram of three protocols, all of which form a circle indicating they are performed iteratively. The protocols are: (i) Baseline protocol is just data collection and validation; (ii) Linguist in the loop (LitL) protocol is Baseline plus expert analyses and guidelines updates for crowdworkers; (iii) LitL Chat protocol is LitL plus a chatroom in which experts and crowdworkers can discuss the task.}
    \includegraphics[width=0.83\linewidth]{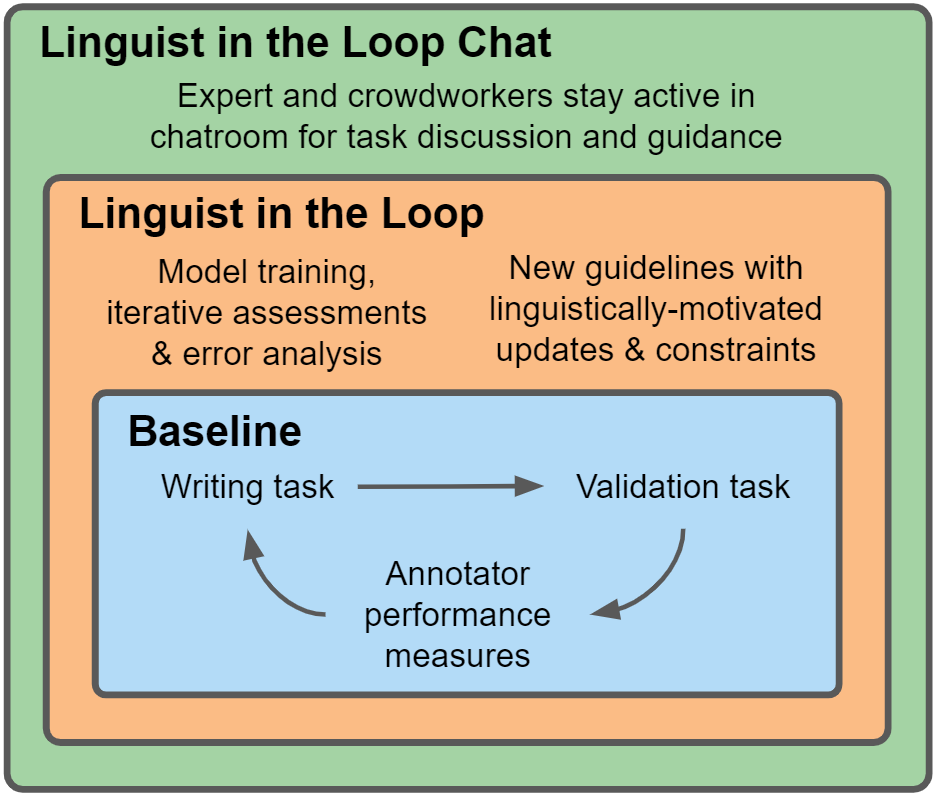}
    \caption{The three protocols compared in this study. Each crowdworker is part of only one protocol.}% The Baseline protocol only includes the two components shown in blue. The Linguist-on-the-Sidelines protocol includes everything from Baseline and additionally all components shown in orange. The Linguist-in-the-Loop protocol includes every component in the diagram.}
    \label{fig:protocol_diagram}
\end{figure}

Many popular datasets for training and evaluating natural language understanding (NLU) models consist of examples written by non-expert annotators. 
While it is convenient and relatively inexpensive to gather large datasets from non-expert crowdworkers, the resulting datasets often suffer from systematic gaps and artifacts. 
Through \emph{post hoc} analysis, experts have identified many such problems and found that augmenting datasets with targeted examples can mitigate these issues \cite{yanaka-etal-2019-help,min-etal-2020-syntactic}. 
Though non-expert created data is often flawed, it is easier to scale up compared to expert annotations, and so it is widely used in the creation of large training datasets. 
With this in mind, we investigate how to leverage expert linguistic knowledge during annotation by having linguists dynamically identify artifacts, biases, and gaps in the data, then communicate with non-expert annotators to instruct them towards annotations that address issues as they arise. % We assess the efficacy of expert intervention through a controlled comparison to a Baseline condition with no direct expert involvement.

We focus on natural language inference \citep[NLI;][i.a.]{dagan2006pascal}, a task where the goal is to predict the label (\textsc{entailment}, \textsc{contradiction}, \textsc{neutral}) that defines the logical relationship of a hypothesis to a premise (e.g., for the premise \textit{Jenny loves all animals} the hypothesis \textit{Jenny loves cats} is an \textsc{entailment}, and \textit{Jenny hates dogs}, a \textsc{contradiction}).
We choose NLI because it is among the best-studied NLU tasks, with demonstrated value, but also multiple well-documented data quality issues that arise in crowdsourced data collection, many of which can be traced to a given \textit{heuristic}. 
%For many data quality issues, a given \textit{heuristic} can be identified that gives insight into the issue's source. 
%Compared to NLI, however, many other NLU tasks have far fewer known heuristics, and many data collection efforts involve new tasks for which we have no known heuristics at the start. 
Because these heuristic-based issues are prevalent, we focus on NLI with the aim that our methodology can inform data collection for new tasks in which there are fewer known heuristics.

%Some NLI benchmarks have attempted to address known issues during data collection, either through limited explicit instructions determined beforehand \citep{hu2020ocnli} or a model-in-the-loop method \citep{nie2020anli}. 
%We expand on these studies by using a linguist-in-the-loop system to introduce expert-guided dataset augmentation into the crowdsourcing process, while avoiding overfitting the data to the biases in a particular model.

Previous attempts to develop more effective NLU data collection protocols have been limited in their ability to assess the efficacy of their interventions, as they often lack direct comparisons between different collection methods. 
%We lay out our assumptions at each step of data collection so that we can explicitly test those assumptions, allowing us to determine the effects of those different design choices.
We test three levels of expert involvement: (i) a baseline group with no hands-on expert involvement (`Baseline'), (ii) a group that followed linguistically-motivated constraints that experts developed to target heuristic-based weaknesses in the data (`linguist-in-the-loop' (LitL)), and (iii) a group that extended the LitL protocol to add direct interaction with the experts, including individual-level discussion about the task on the chat platform Slack (`LitL Chat'). 
These three protocols are shown in Figure~\ref{fig:protocol_diagram}, and a task example with a constraint from LitL and LitL Chat is shown in Figure \ref{fig:sample_hit}.%The role of linguists in our study is to iteratively provide additional guidelines and instructions on how to appropriately modify the original texts and answer questions that are raised during the process. This may potentially change or rule out certain heuristics that would otherwise endure.

% \begin{figure}[t]
%     \centering
%     \includegraphics[width=0.9\linewidth]{figs/HIT_one-col_no-border.png}
%     \caption{Round 5 HIT with the optional \textit{No Overlap} constraint shown.}
%     \label{fig:sample_hit}
% \end{figure}

\begin{figure}[t]
\centering
\begin{minipage}{0.92\linewidth}
\newcommand{\exspacing}{0.5}
\newcommand{\exspacingb}{0.9}
\tiny

\textbf{\scriptsize Text:} They inhabit the near-boiling water of geysers in Yellowstone, and the even hotter water in volcanic vents on the ocean floor. \newline

$\chkbox$ The definitely correct sentence {\color{red}does not reuse nouns, verbs, adjectives, or adverbs from the text} (\$0.10) \vadjust{\vspace{\exspacing pt}}\newline
$\bullet$ Your definitely correct statement must not contain any of the following words: {\color{red}there, can, may, might, some, people} \vadjust{\vspace{\exspacingb pt}}\newline
\textbf{\scriptsize Definitely correct statement:}\vadjust{\vspace{\exspacing pt}}\newline
\framebox[\linewidth]{}\vadjust{\vspace{3pt}}\newline

$\chkbox$ The maybe correct sentence {\color{red}does not reuse nouns, verbs, adjectives, or adverbs from the text} (\$0.05) \vadjust{\vspace{\exspacing pt}}\newline
$\bullet$ Your maybe correct statement must not contain any of the following words: {\color{red}often, several, many, most, some, other, will} \vadjust{\vspace{\exspacingb pt}}\newline
\textbf{\scriptsize Maybe correct statement:}\vadjust{\vspace{\exspacing pt}}\newline
\framebox[\linewidth]{}\vadjust{\vspace{3pt}}\newline

$\chkbox$ The definitely incorrect sentence {\color{red}does not reuse nouns, verbs, adjectives, or adverbs from the text} (\$0.05) \vadjust{\vspace{\exspacing pt}}\newline
$\bullet$ Your definitely incorrect statement must not contain any of the following words: {\color{red}any, never, no, nothing, not/n't, only, always, all} \vadjust{\vspace{\exspacingb pt}}\newline
\textbf{\scriptsize Definitely incorrect statement:}\vadjust{\vspace{\exspacing pt}}\newline
\framebox[\linewidth]{}

\end{minipage}

\caption{Round 5 HIT with the optional \textit{No Overlap} constraint shown.}
\label{fig:sample_hit}
\end{figure}

We observe that expert involvement (LitL and LitL Chat) during data collection reduces the impact of some learnable heuristics and results in a more challenging final dataset with model performances that are 5 points lower on validated data compared to Baseline. 
%We observe that iteratively-updated expert guidelines during data collection reduce the impact of some learnable heuristics and result in a more challenging final dataset with model performances that are 5 points lower on validated data compared to Baseline. 
%We evaluate model performance using validated data to make sure lower performance is not simply the result of noise. 
Examples in all protocols qualitatively appear equally free of noise (incorrect labels, typos, etc.), and lexical diversity increases in later rounds for the protocols with linguist intervention.\footnote{Appendix \ref{app:examples} contains a sample of validated examples.} 
However, we find no evidence of better model accuracy on adversarial examples or out-of-domain datasets. 
Further, we do not find any benefit to providing a chatroom for crowdworkers to interact directly with linguists. %, beyond the benefits of simply updating targeted guidelines. 

% In order to examine the precise effect of this newly introduced intervention, we design two alternative loops in which the linguists are either less actively involved or not involved at all.

\section{Related Work} %Alicia, Nikita

% NLI datasets & their annotations
\paragraph{NLI Data Collection Methods}
Large-scale human-elicited datasets include the Stanford Natural Language Inference Corpus \citep[SNLI;][]{bowman2015large}, the Multi-genre Natural Language Inference Corpus \citep[MNLI;][]{williams2020mnli}, the Chinese OCNLI corpus \citep[][]{hu2020ocnli}, and Adversarial NLI \citep[ANLI;][]{nie2020anli}. 
All four datasets use non-expert annotators to write hypotheses and annotate inference labels from pre-defined short texts, though only OCNLI and ANLI add interventions to increase data diversity. 
OCNLI %adopts a similar design to MNLI, but 
hires students specializing in linguistics and language studies to construct hypotheses and adds data collection rounds with instructions for avoiding known sources of bias. %, focusing primarily on reducing the effect of certain lexical cues. % hypothesis-only bias. %, i.e., corpus level cues in hypotheses that make it possible to achieve better-than-chance performance without using the premises at all.
ANLI uses a human-and-model-in-the-loop procedure to elicit examples that are progressively more difficult for their model, resulting in a dataset with a large human--model performance gap, 
%We build on these existing datasets by adopting the same task format, but testing the degree to which expert guidance shapes the resulting dataset.
%The benefit of building a human-annotated corpus for a language other than English has been attested using the Original Chinese Natural Language Inference dataset (OCNLI) \citep{hu2020ocnli}. OCNLI adopts the basic design of MNLI and employs experts specializing in linguistics and language studies to construct hypotheses. %The results reassure the overall accomplishment of human expert annotation. 
%Recently, human-and-model-in-the-loop processes like those used in ANLI (and `Beat the AI' \cite{bartolo2020beat} in Question Answering) have created challenging benchmarks for state-of-the-art models by focusing on examples on which a pretrained model fails. 
though identifying the cause for model failure is left up to the discretion of the worker. 

%\paragraph{Experimental Comparisons of Data Collection Methods}
Efforts to improve on sentence writing tasks for NLI have yielded mostly negative results in head-to-head protocol comparisons. 
In an experimental comparison on different NLI crowdsourcing protocols, \citet{vania2020asking} find that automatically selecting premise-hypothesis pairs % via (i) a similarity comparison metric, and (ii) translation.
for crowdworker annotation does not yield a better dataset compared to a baseline sentence writing protocol.
\citet{bowman2020new} compare interventions aimed at improving NLI writing, using protocol variants that constrain the worker's task, but they do not see improvements in transfer learning results %have crowdworkers make small edits to existing premise-hypothesis pairs, use longer premises, or write hypotheses based on a contrasting pair of premises. 
% However, they find that none of the interventions improve the resulting dataset for transfer learning 
compared to a baseline protocol. %which is effectively identical to ours.

\paragraph{NLI Dataset Bias}
Several studies have identified systematic biases in NLI datasets that the models trained on them subsequently learn (often robustly). 
Hypothesis-only bias, where a model correctly labels the relationship of a premise-hypothesis pair based only on the hypothesis, is a well-documented issue \citep[i.a.]{poliak-etal-2018-hypothesis,gururangan2018artifactsNLI}. %, for which there are already several proposed solutions including ... 
Lexical overlap between the premise and hypothesis is another source of bias \cite{mccoy-etal-2019-right,naik2018stress}, as greater overlap between a premise and hypothesis is associated with a greater likelihood the label is \textsc{entailment}. %{\color{red}others}, 
Sources of bias can also be due to gaps in the training data, and \citet{sinha2020unnatural} point to the lack of syntactic understanding in NLI models as one such example, noting that models often ignore syntactic information entirely.
The well-studied biases in NLI make the task a good test case for protocols designed to assess biases as data is collected.

\paragraph{Methods for Filling the Gaps in Datasets}
To collect challenging examples for NLU tasks, researchers have explored altering labeled data to create targeted or adversarial examples. 
\citet{kaushik2020learning} have crowdworkers make minimal edits to annotations to align with a revised label. 
\citet{gardner2020evaluating} create contrast sets for evaluation by having experts alter already-annotated examples such that the resulting label changes. 
\citet{wei-zou-2019-eda} use simple automatic data manipulations to augment datasets for several text classification tasks, resulting in more robust models. 
More linguistically sophisticated manipulations have been used to augment MNLI to improve monotonicity reasoning \cite{yanaka-etal-2019-help} and mitigate lexical overlap heuristic \cite{min-etal-2020-syntactic}. 
%They argue that non-experts (here, anyone not involved in the dataset's creation) would lack an understanding of which examples to perturb and in what ways, solidifying the need for expert annotations in dataset adjustment.
These methods are applied after crowdsourced data collection is complete, so it is not clear if the gaps they identify in a final dataset would have been avoidable if addressed during the data collection process.

Most similar to our approach, OCNLI's instructions nudge annotators towards writing examples that address \textit{known} sources of bias.
%Though they do not assess the degree to which annotators followed these suggestions, 
They find that encouraging annotators to follow constraints such as avoiding negation in a \textsc{contradiction} label results in a harder dataset. 
We expand on this work by introducing a wider range of constraints and assessing their effects throughout data collection. 
% beyond NLI datasets for addressing dataset gaps and biases
Our approach is also similar to \citeauthor{vidgen2020learning}'s (\citeyear{vidgen2020learning}) human-generated hate-speech dataset. 
They introduce \textit{pivots} during data collection in which they instruct annotators about how to write to fool their model.
We expand on their method by %not only drawing inspiration from known sources of bias, but also dynamically 
qualitatively assessing the annotations to identify issues specific to our data as it is collected.

\paragraph{Expert Interaction with Crowdworkers}
% reported benefit to direct communication with workers
\citet{DBLP:conf/www/TangYH19} report that direct communication among crowdworkers leads to improved task performance on image labeling, optical character recognition, and audio transcription. 
This suggests that collecting higher quality data is possible when workers have real-time group interaction.
Other studies have reported that interaction among crowdworkers is an effective tool for limiting some forms of bias and increasing accuracy \citep{drapeau2016microtalk, schaekermann2018resolvable}.
\citet{roit2020controlled}, in a different strategy, gives annotators detailed feedback during training, then selects only a small number of those workers for the larger annotation task.
This strategy frontloads the work of the experts and relies on the selected workers to perform the task consistently.

% recap the gap that this project fills
%We ask whether the benefit of targeted interventions highlighted in previous studies can be dynamically shaped by expert linguists into crowdworker-friendly guidelines.
Despite the potential benefits of real-time interaction between crowdworkers and experts, there has not yet been a direct comparison of protocols that differ based on this variable. 
%We investigate the degree of expert involvement needed for such an intervention to be successful.
To our knowledge, this study is both the first to test the effect of this interaction and the first head-to-head experimental assessment of human-in-the-loop dataset collection methods, allowing us to make conclusions about the causal effects of the different interventions compared to a baseline.

\section{Dataset Collection} 
\label{subsec:datacollec}

\paragraph{Task Description}
%We conduct five one-week-long rounds of data collection with protocols
Our task is modeled on MNLI's data collection procedure.
We present workers with a text, for which they write statements they consider definitely correct, maybe correct, and definitely incorrect. % (corresponding to entailment, neutral, and contradiction).
Following each round of sentence writing, crowdworkers validate 500 examples from their protocol. % (14\% of the data).
%Workers only validate writing from the protocol they were assigned to.
%In the validation HITs, workers label six premise-hypothesis pairs.
We collect four validations for each example that we validate and use these labels plus the original one to assign a gold label based on majority vote.
Examples for which no gold label can be assigned are removed from the data.
We use the validated data to evaluate our models and the unvalidated data for training.
Workers with a validation rate below 70\% or whose validation responses fail to match the gold label at least 70\% of the time are subject to disqualification. 
Throughout the study, we disqualified three workers from Baseline, three from LitL, and two from LitL Chat.

\paragraph{Pay Structure}
To retain crowdworkers for all five rounds, we increase the base pay of \$1/HIT by \$0.05 each round and pay a \$20.00 bonus after the last round. 
To ensure we collect sufficient examples from each worker, we award a bonus worth 10\% of their base pay for reaching milestones of 10, 50, and 100 HITs each round. % within a round. %\footnote{If 1 HIT is worth \$1.00 as in Round 1, a worker receives \$1.00 bonus for completing 10 HITs, \$5.00 bonus for completing 50 HITs, and \$10.00 bonus for completing 100 HITs.}
To encourage workers to write high-quality examples, we pay a \$5.00 bonus each round to workers with over 25 HITs and at least a 95\% validation rate.
We estimate that, with bonuses, a worker who completes 70 HITs with a high validation rate will earn \$81 in Round 1 ($\sim$\$16/hr), with that rate rising to \$95 in Round 5 ($\sim$\$19/hr).
Workers in LitL and LitL Chat earn additional bonuses for completing challenge options (\$0.05-\$0.10), and workers in LitL Chat earn bonuses for participation in the chatroom (\$1.50 for any engagement, \$10.00 for active engagement).

\subsection{Crowdworker recruitment} 
We use a pre-test to recruit workers via Amazon Mechanical Turk (MTurk).
The pre-test is open to workers in the United States with approval rates at or above 98\% and more than 1000 HITs approved.
The pre-test is a sentence-writing task where workers see a premise and write %sentences about it that are definitely correct (\textsc{entailment}), maybe correct (\textsc{neutral}), and definitely incorrect (\textsc{contradiction}).
hypotheses under each of the three NLI labels.
To assess if workers can follow more complicated instructions, they are also asked to write one entailed sentence that uses a conjunction and one neutral sentence that does not re-use any words from the text.

We collect responses from 155 crowdworkers, of whom 145 indicate interest in completing future, similar HITs. 
From those 145, we read their responses and exclude 24 for failing to adequately complete the task (many due to responses that do not follow instructions).
The remaining 121 crowdworkers are retained and split between the three experimental protocols in a pseudo-random way such that (i) the three workers who asked not to participate in a chat forum are placed in the Baseline or LitL protocol,\footnote{Though a potential design confound, this was necessary and had minimal effect. \textit{Requiring} workers to sign up for a third party service violates Amazon's terms of service, so we allow participants to opt out. Only three participants opted out of the chat (two of whom dropped out after Round 1), and many workers placed in a non-chat protocol had indicated a willingness to participate in the chat.} and (ii) groups are matched based on a 4-point rating scale of their qualitative performance on the pre-test. 
We ultimately had 37 annotators participate in data collection in Baseline, 30 in LitL, and 32 in LitL Chat.

\subsection{Annotation Details} % Soo-Hwan, Alex
\label{sec:annot_details}

Crowdworkers annotate data in five rounds, with each round lasting one week. 
Between rounds, we conduct several planned diagnostics on our datasets to monitor the impact of our intervention and inform annotator feedback for the following round.

\paragraph{Annotation stage}
Annotators construct hypotheses based on premises taken from the \textsc{Slate} subset of MNLI. \textsc{Slate} hosts popular culture articles from the archives of Slate Magazine. %\footnote{We exclude other genres for having short premises, covering tedious topics, or lacking sufficient structural variation.}
%Among the genres available in MNLI, we used the premises in one particular genre, namely Slate.
After Round~1, we exclude premises that are shorter than six tokens based on feedback from annotators that many of the very short premises are incomplete, nonsensical, or confusing to write hypotheses for.

\paragraph{Diagnostic stage}
% After each round of data collection, we perform several planned diagnostics on our datasets to monitor the impact of our intervention and inform our feedback for the following round.
After each round, we fine-tune RoBERTa \citep{liu2019roberta} models using data collected up to that round. 
We then evaluate the models on diagnostic examples from GLUE \citep{wang2019GLUE} and HANS \citep{mccoy-etal-2019-right}.
The GLUE examples target different aspects of linguistic reasoning including lexical semantics, predicate-argument structure, logic, and world knowledge.
HANS tests for three shallow heuristics, including lexical overlap between a premise and hypothesis.
We also train and evaluate RoBERTa models using hypothesis-only inputs to assess hypothesis-only biases in the data \citep{gururangan2018artifactsNLI}. 
Finally, we assess the distribution of hypothesis lengths and the pointwise mutual information (PMI) between each word in the vocabulary and label.
Hypothesis length does not appear to differ by protocol or label, so it never informs our constraints. %; we do not discuss this metric further.

We use these diagnostics as well as qualitative reviews of the data to devise linguistically-motivated guidelines for the following round, allowing us to adapt feedback for crowdworkers in a structured way as the data is collected. 
This process is conducted by five of the authors who have graduate training in English syntax and semantics.

\begin{table*}[ht]
    \centering
    \resizebox{\textwidth}{!}{% 
    \rowcolors{2}{gray!25}{white}
    \begin{tabular}{p{0.125\textwidth}p{0.42\textwidth}p{0.42\textwidth}rrr}
    \toprule
    \bf Constraint&\bf Premise&\bf Hypothesis&\bf Label&\multicolumn{2}{c}{\bf Attempt rate} \\
    & & & & LitL & Chat\\\midrule
Hypernym or hyponym & Does anyone know what happened to \textbf{chaos}? & Whatever happened to the \textbf{lack of order} is certainly a mystery. & E & 22.8 & 23.7\\
Banned word in diff. label & Inflation is supposed to be a deadly poison, not a useful medicine. & \textbf{All} people believe inflation is supposed to be a useful medicine & C & 43.7 & 27.7\\
Temporal reasoning & John Kasich dropped his presidential bid. & They said that \textbf{earlier}, John Kasich had dropped his presidential bid. & E & 34.1 & 10.0\\
Synonym or antonym & 2) This particular instance of it \textbf{stinks}. & This instance is perceived to be \textbf{a good thing}. & C & 39.5 & 24.5\\
%Hyponym&But still Wendy Wasserstein eludes us.&Windy Wasserstein confounds us.&E\\
All overlap & News argues that most of America's 93 million volunteers aren't doing much good. & News argues that volunteers aren't doing much good. & E & 21.8 & 30.4\\
Register change & First, the horsemen brought out a teaser horse. & Teaser horses are commonly thought to be both entertaining and tragic. & N &25.3 & 15.0\\
%Antonym&It's a full-time job just controlling the young hotheads on some NBA squads.&The NBA is devoid of young stable men.&E\\
No overlap & and she doesn't floss while driving. & The woman has an automated car. & N &29.2 & 22.3\\
Relative clause & Sun Ra's spaceships did not come, as it were, out of nowhere. & The spaceships \textbf{that belong to Sun Ra} came out of nowhere & C & 35.0 & 24.3 \\
Reverse argument order & After an \textbf{inquiry} regarding \textbf{Bob Dole}'s ... & It is illegal for \textbf{Bob Dole} to receive \textbf{inquiries}. & N &36.7 & 29.4 \\
Grammar change & The Bush campaign \textbf{has} a sweet monopoly on that. & The Obama campaign \textbf{had} a sweet monopoly on that. & C & 22.6 & 13.4 \\
Sub-part & He was crying like his mother had just walloped \textbf{him}. & He cried a lot, as though he were walloped on \textbf{his behind}. & E & 23.2 & 19.1 \\
Background knowledge & In both \textbf{Britain and America}, the term covers nearly everybody. & The term generally applied to \textbf{countries in two opposite sides of the world}. & E & 32.9 & 15.9 \\\bottomrule
    \end{tabular}
    }
    \caption{Sentence pairs displaying each challenge option. Where applicable, relevant contrasts are bolded. Examples are randomly drawn from data that passed validation on the constraint with the restriction that both sentences be fewer than 80 characters ($\sim$ 32\% of the data). % and both the original annotator and at least two validators agree that the example satisfies the constraint. 
    The last column shows the percentage of the challenges attempted.}
    \label{tab:guideline_examples}
\end{table*}

\subsection{Constraints}
\paragraph{Banned Words} 
After Round 1, crowdworkers in LitL and LitL Chat are instructed not to use certain words when writing sentences for each label. 
%The banned words are determined by selecting a set of words with the highest PMI in the LitL and LitL Chat protocols.
In each round, 5-7 new banned words are identified based on the PMI for each label. 
We use PMI to identify words to ban under each label, as words with high label PMI are a major contributor to hypothesis-only bias. 
This constraint is mandatory in all HITs. 
% to counter any bias from differences in lexical distribution across labels. 
Figure \ref{fig:sample_hit} shows examples of the banned words during Round 5.
%The banned words for Round 5 are as follows: \textit{there}, \textit{can}, \textit{may}, \textit{might}, \textit{some}, and \textit{people} for entailment, \textit{many}, \textit{most}, \textit{some}, \textit{other}, \textit{will}, \textit{would}, and \textit{often} for neutral, and \textit{no}, \textit{never}, \textit{not/n't}, \textit{only}, \textit{always}, \textit{all}, and \textit{any} for contradiction. 

\paragraph{Challenge Options} 
We use constraints, framed as \textit{challenge options} to the worker, to target heuristics that we identify in the data during the diagnostic state. 
By explicitly telling workers to avoid these heuristics, we aim to lower their contribution to any bias in the final dataset. 
We determine constraints through qualitative assessment of the data, taking into consideration syntactic diversity, lexical choice, and semantic or world-knowledge-based reasoning patterns.
For example, after noticing that the majority of hypotheses relied only on the stated information from the premise in Round 1, we encouraged workers in Round 2 to focus on ``background knowledge'' (example in Table \ref{tab:examples}) that they know to be true, but isn't explicitly stated, such as the knowledge that Britain and America are countries on opposite sides of the world. 

After Round 1, each HIT in LitL and LitL Chat lists one constraint. 
This task is optional for the workers, as some constraints are incompatible with some examples.
%We leave it up to the workers' best judgment when it makes sense to apply the constraint, with the assumption that this will lead to fewer unintended artifacts in the data brought on by trying to fit each example to a given constraint. 
%Each HIT offers only one optional guideline, and we release HITs sequentially in batches all offering the same guideline. In each round, we offer different guidelines. 
%The guidelines can be conceptually grouped into three major categories: lexical choice, syntax, and world knowledge. 
%The choice of constraints to offer as challenge options for the workers in each round is determined from a mix of reviewing the results of the iterative analyses and reading through the validated examples to look for patterns in responses.
%A reason for asserting these linguistically-motivated guidelines is to induce an appropriate amount of variation in data. In the end, we expect them to mitigate the overrepresentation of a particular aspect of language. 
The 12 challenge options are defined in Appendix \ref{challenge_options}, % and grouped as either lexical choice, syntax, or world knowledge. 
with examples in Table \ref{tab:guideline_examples}.

\subsection{Protocols}
\label{subsec:protocols}

\paragraph{Baseline Protocol}
Our Baseline protocol follows the basic task description in the beginning of \S \ref{subsec:datacollec} and does not include any direct expert involvement.

%As a baseline, we crowdsource NLI examples by providing non-linguistically-motivated feedback between iterations. We use this type of feedback primarily to compare it with other types of non-baseline conditions. 

%We include a bonus structure in which crowdworkers are rewarded at the end of each round.
%. Weekly bonus features are offered to each annotator based on how many HITs they complete (10, 50, or 100 HITs) and how accurate their HITs are (validation of 95 percent or above for 25+ HITs). A final bonus feature is awarded at the end of the study if an annotator submits at least one HIT during each of the five planned weeks.

\paragraph{Linguist-in-the-Loop (LitL) Protocol}
LitL extends the Baseline protocol with constraints (described in \S \ref{sec:annot_details}).
As the constraints make the task more difficult, we award bonuses to workers who indicate that they attempted a challenge option.
The bonus is \$0.05-\$0.10 per example, determined by the linguists' assessment of the difficulty. 
For example, the \textit{No Overlap} constraint is more difficult in entailment examples than neutral, so a \textit{No Overlap} entailment example receives a higher bonus.

During the validation round, examples with challenge options are validated for whether they adhere to the given constraint, regardless of whether the annotator indicated that they had attempted the challenge. 
For any worker whose validation rate on the bonus challenges is below 50\%, we contact them to explain the source of their errors. % and ask them to read the instructions more closely in the future.
%We did not remove any workers from the experiment due to poor performance on the challenge options.
%We check examples approximately once every hour to make sure that workers are following the banned words constraint, as this process can be semi-automated.
%Workers are given feedback via email on the specific examples that do not follow the constraint, and are allowed to continue working on HITs only after acknowledging the feedback.

%Indirect linguistic feedback is provided between collection iterations when necessary. This involves either encouraging the crowdworkers or banning them from participating in future tasks. The decision largely depends on the annotator's accuracy in performance. 

%The bonus system is the same as the one in the baseline condition except that there is an added feature which rewards annotators based on how many heuristics they rule out for each HIT. This feature is not included in the first week of the data collection phase.  
%There may exist ongoing training, oversight, and indirect contact with experts.

\paragraph{LitL Chat Protocol}
%Our LitL+chat protocol extends the procedure for the LitL protocol.
We provide direct communication with expert linguists on Slack.
%The primary function of Slack was to answer questions as they arise.
We encourage workers to ask task-specific questions for anything they find challenging or confusing.
Most questions seek to clarify if a certain strategy `counts' as adhering to a constraint.
%Each round, HITs are released in 7-9 batches, and a linguist is on call in the forum for the two hours following each batch release.
Feedback given via email in the LitL protocol is instead given via direct message in Slack, unless the worker initiates contact over email.
Additionally, at the beginning of Rounds 3--5, we identify creative examples written in a previous round and post them to Slack for inspiration, with a brief comment.

%Workers are allowed to discuss their questions with each other and offer their perspective and advice.
%Our live feedback consists of giving strategy-based suggestions, but not word-for-word answers to what workers should write.

%This active interaction is expected to mitigate or rule out certain heuristics identified in the current dataset. 

%To compensate workers for time spent on Slack, we pay a \$1.50 bonus at the end of each round to workers who have interacted via Slack at all, defined as having made at least one post or reply.
%To encourage active and helpful participation, we pay \$10.00 to workers who meet a higher standard of substantially contributing to the Slack discussion.
%Active and substantial interaction was determined qualitatively by reviewing the posts by each worker. 

%\subsection{Linguist Involvement}
%We use three levels of expert involvement while iteratively collecting NLI data to evaluate how best to use experts during crowdsourcing. In each level, crowdworkers are asked to modify various texts collected from MNLI \citep{williams2020mnli} by performing a 3-class classification task (entailment, contradiction, and neutral) as a prequalification. We then assign crowdworkers into one of three groups that vary by the type of feedback they receive during and between collection iterations.

\subsection{Annotator Performance}

\paragraph{Inter-Annotator Agreement} 
%Recent work indicates that there is substantial variation in human judgements for NLI tasks \citep{pavlick2019inherent, min2020ambigqa, nie2020can, williams2020anlizing}. 
%We calculate Fleiss' Kappa within each protocol to assess the rate of inter-annotator agreement; as a rule of thumb, Kappa above 0.6 can be considered `substantial agreement'.
Baseline shows the highest inter-annotator agreement with $\kappa$ = 0.71, while LitL and LitL Chat have 0.64 and 0.63, respectively. 
All three protocols meet the standard threshold for ``substantial agreement.''
%Our tasks also require human judgements, which suggests that some amount of variation is expected. 
% We also assess validation rates in each protocol.
%As a way of analyzing variation, we measure inter-annotator agreement across all three protocols. In the validation rounds, each annotator is asked to validate sentences constructed by other annotators. The gold label for each sentence is based on the majority vote from five annotators. 
Validation rates are 93.7\% for Baseline, 89.76\% for LitL, and 91.36\% for LitL Chat. 
LitL and LitL Chat may have slightly lower validation rates than Baseline because the constraints lead to challenging examples, making the validator's task more difficult. % the validation task is affected by the difficulty of the annotator's task. %the annotator's task is harder, resulting in more challenging examples. 
%In general, however, the average validation rates are reasonable in all protocols.      

\paragraph{Proportion of Constraints Attempted}
%In Rounds 2-5, crowdworkers in LitL and LitL Chat are given the option to do bonus challenges. 
%They were allowed to fill in a checkbox each time they targeted a single heuristic. 
The attempt rate of bonus challenges differs substantially between constraints (Table \ref{tab:guideline_examples}). 
Overall, more abstract categories (e.g., background knowledge) are attempted less often than more concrete constraints. %, possibly reflecting different difficulty levels for different constraints.
There are also differences by protocol, as LitL had a higher attempt rate than LitL Chat. 
One reason for this difference may be that workers in LitL Chat were more selective about identifying good examples on which to apply the constraints.
Supporting this possibility, we find that LitL Chat had higher constraint validation rates than LitL in Rounds 4 and 5, indicating that workers in LitL Chat adhered to the constraints more accurately after practice. 
%In rounds 2 and 3, the validation rate on the constraints is similar between the two protocols, but in rounds 4 and 5, workers in LitL Chat achieved higher constraint validation rates.
%This difference indicates that workers in LitL Chat adhered to the constraints more accurately after practice, and so the higher attempt rate in LitL corresponds to an effective increase in examples that follow the constraints of only about 20\% for LitL over LitL Chat. 
% .627*110 lots vs. .789*76 litl (round 4)
% = 68.97           59.9 --> 15% more
% .713*143 lots vs. .882*93 litl (round 5)
% = 101.95          82.02 --> 24% more
%Although workers found the task-related instructions clear, some reported that providing more monetary bonuses may have led to a higher participation rate.

\paragraph{Use of Slack}
%All crowdworkers in LitL+chat are encouraged to communicate with linguists through a Slack channel, though some individuals choose not to be active in the discussion.
The total number of active workers in Slack fell from 23 in Round 1 to just 16 by Round 4.\footnote{Round 5 was even lower, but spanned the US Thanksgiving holiday, which likely artificially lowered participation.}
The total number of messages sent also fell with each round, going from about 215 posts and replies in Round 1 to 162 in Round 4.
It may be that workers rely on the chat less as they become more familiar with the task. 
%Many of the workers participated actively by asking questions about certain tasks. Some of them even replied to others' questions by suggesting appropriate ways of solving the tasks. We plot Slack usage over time in Figure \ref{fig:slack}. Usage is highest on the days when HITs are released and declines significantly over the five rounds.

%\begin{figure*}
%    \centering
%    \includegraphics[width=\textwidth]{figs/slack_messages.png}
%    \caption{Number of messages sent over the Slack channel (LitL group only) over time.}
%    \label{fig:slack}
%\end{figure*}

\section{Experiments} % will, emily, karmanya
For each round and protocol, we collect 3.5k examples and use the 500 validated examples (\S \ref{subsec:protocols}) as validation data and the remaining 3k for training.\footnote{Data and code: \href{https://github.com/Alicia-Parrish/ling_in_loop/tree/master}{github.com/Alicia-Parrish/ling\_in\_loop}} We then fine-tune a RoBERTa$_{\rm{Lg}}$ model on all data accumulated up to Round $n$% to study corpus characteristics of a dataset collected using $n$ iterations of our protocol
. For example, the Round 2 model is trained on examples from Rounds 1 and 2 with training and validation sizes of 6k and 1k, respectively. We also fine-tune a RoBERTa$_{\rm{Lg}}$ model previously trained on MNLI (RoBERTa$_{\rm{Lg+MNLI}}$) and find similar results (details in Appendix \ref{app:mnli_trained_results}). After each round, we evaluate our models on the diagnostics described in \S \ref{sec:annot_details}. %, which we refer to as iterative diagnostics. 

After the final round of data collection, we evaluate models trained on our data on MNLI-mismatched \citep{williams2020mnli} and ANLI \citep{nie2020anli}. 
The MNLI corpus includes two evaluation sets,  MNLI-matched and MNLI-mismatched, with examples sourced from different genres. 
We evaluate on MNLI-mismatched, as we source our premise sentences from an MNLI-matched genre. 
Evaluating on held-out sets allows us to test if our interventions lead to increased model accuracy on datasets generated through different protocols or from different sources while ensuring that we do not overly tune our feedback to these benchmarks.

We estimate average accuracy and confidence intervals by fine-tuning 10 additional models with a sample of 90\% of the collected training data. We use the best hyperparameters for each protocol and round from our hyperparameter search described below. In sampling the data, we first sort the data by annotator and successively remove 10\% of examples, allowing us to study variation among annotators while controlling for training set size.

\paragraph{Implementation}
To fine-tune our models, we perform a grid search over learning rate $ \in \{5e-6, 1e-5, 2e-5, 3e-5\}$ and batch size $ \in \{16, 32\} $ and use the hyperparameters yielding the best in-domain validation accuracy. We train for 20 epochs, since each round of data collection yields 3k training examples, and longer training has been shown to help smaller training sets \citep{Zhang20Revisiting}. Our code %\footnote{\href{https://github.com/Alicia-Parrish/ling_in_loop/tree/master}{https://github.com/Alicia-Parrish/ling\_in\_loop}} 
is based on \texttt{jiant} \citep{phang2020jiant}, which uses PyTorch \citep{pytorch2019} and Transformers \citep{wolf-etal-2020-transformers}.

\section{Results} % Will, Emily, karmanya

\begin{figure}[h]
    \centering
    \includegraphics[width=0.5\textwidth]{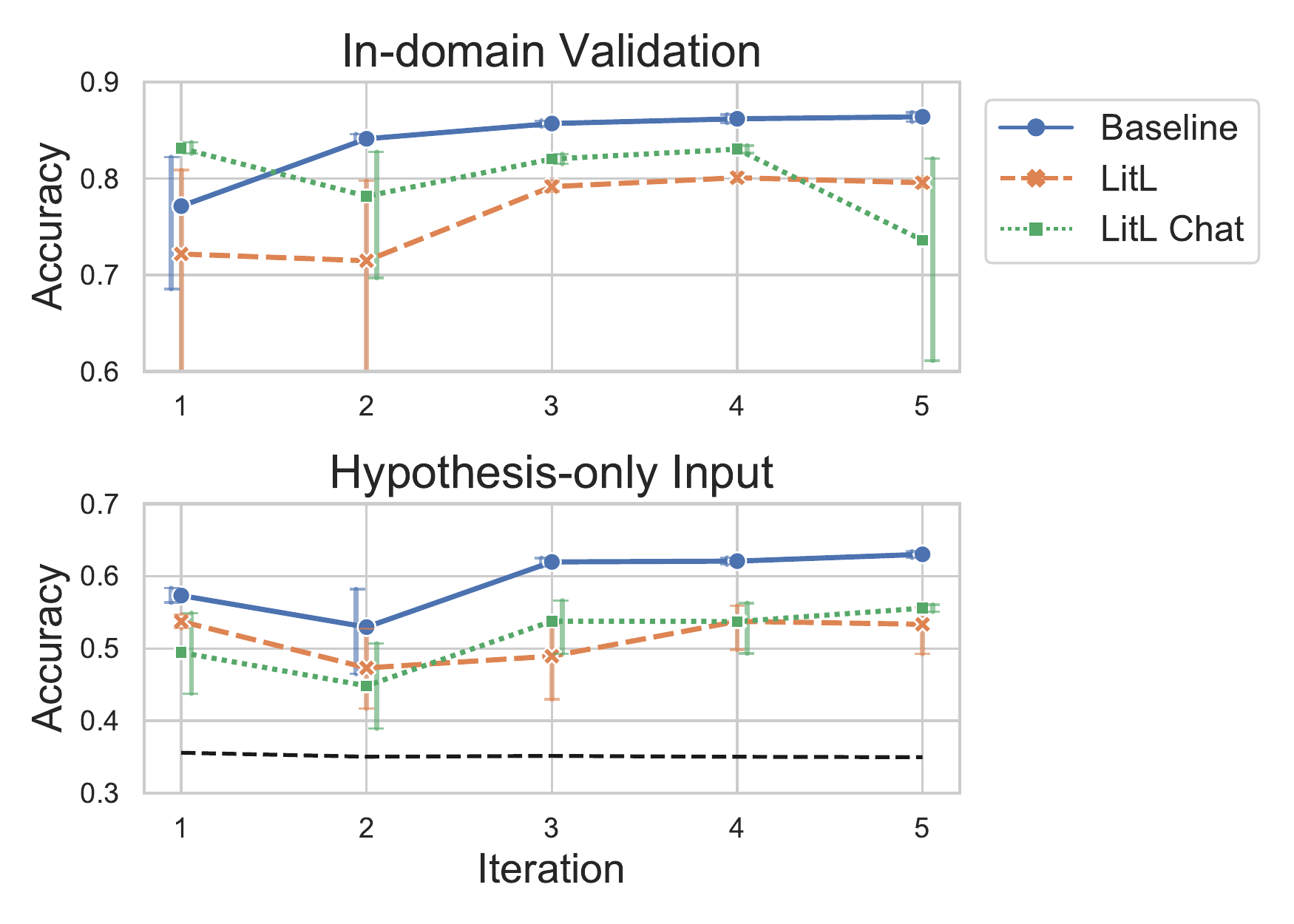}
    \caption{Performance of RoBERTa$_{\rm{Lg}}$ fine-tuned on data collected through different protocols on in-domain validation data trained with either the normal sentence pairs (top) or hypothesis-only (bottom) input. Higher hypothesis-only accuracy indicates more bias. For each round, we include training and validation data \textit{accumulated} up to Round $n$. Dashed black line marks average majority class baseline across protocols.}
    \label{fig:indomain_val}
\end{figure}

\begin{figure*}
    \centering
    \includegraphics[width=\textwidth]{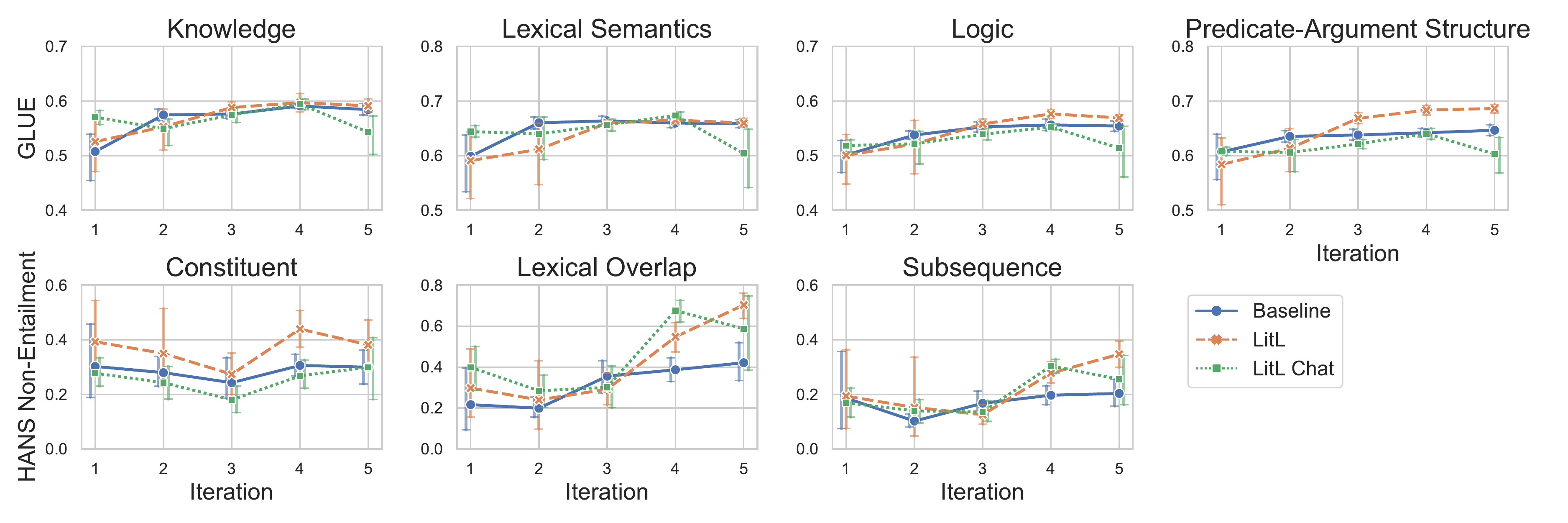}
    \caption{Performance of RoBERTa$_{\rm{Lg}}$ fine-tuned on data collected through different protocols on the GLUE diagnostic set (top) and HANS non-entailment examples (bottom).}
    \label{fig:iter_val}
\end{figure*}

\begin{table}[]
    \centering
    \small
    \begin{tabular}{l r r r}
        \toprule
        \bf Round & \bf Baseline & \bf LitL &\bf LitL Chat \\
        \midrule
        R1 & 84$_{\text{(1.2)}}$ & 82$_{\text{(1.1)}}$ & 86$_{\text{(1.1)}}$ \\
        R2 & 85$_{\text{(0.6)}}$ & 81$_{\text{(1.0)}}$ & 85$_{\text{(0.9)}}$ \\
        R3 & 86$_{\text{(0.7)}}$ & 82$_{\text{(0.8)}}$ & 84$_{\text{(0.9)}}$ \\
        R4 & 86$_{\text{(0.8)}}$ & 83$_{\text{(0.7)}}$ & 84$_{\text{(1.0)}}$ \\
        R5 & 87$_{\text{(0.5)}}$ & 82$_{\text{(0.6)}}$ & 84$_{\text{(0.9)}}$ \\
        \bottomrule
    \end{tabular}
    \caption{Average performance of RoBERTa$_{\rm{Lg}}$ fine-tuned on MNLI over 10 random restarts on validated examples accumulated up to Round $n$. Values in parentheses indicate standard deviation of performance.}
    \label{tab:mnli-only_indomain}
\end{table}

\paragraph{Evaluation Difficulty}
We test whether data collected with expert intervention leads to a more challenging test set by comparing  in-domain performance for each protocol for RoBERTa$_{\rm{Lg}}$, %\footnote{See Appendix \ref{app:mnli_trained_results} for all RoBERTa$_{\rm{Lg+MNLI}}$ results.} 
using training and validated evaluation data accumulated up to Round $n$ (Figure \ref{fig:indomain_val}). 
This allows us to study the characteristics of an iteratively collected corpus using $n$ rounds of each protocol. 
We see that LitL and LitL Chat performance falls below Baseline after the introduction of linguistically-informed constraints in Round 2. % (two-way ANOVA of rounds and protocols between Round 1 and 5 result in p = 2.3e-2 and 5.6e-20 RoBERTa$_{\rm{Lg}}$ and RoBERTa$_{\rm{Lg+MNLI}}$, respectively)
Table \ref{tab:mnli-only_indomain} shows a similar trend -- performance from RoBERTa$_{\rm{Lg}}$ fine-tuned \textbf{only on MNLI} on the same validation sets % as in Figure \ref{fig:combined_val}
decreases or remains relatively low for LitL and LitL Chat, while performance on Baseline increases as more data is collected. 
As we evaluate on validated examples, it is unlikely that this lower performance is due to noise in the data. Rather, these findings indicate we are able to create more challenging evaluation data using the LitL and LitL Chat interventions.

\paragraph{Hypothesis-Only Bias}
We test whether the data collected with linguist intervention leads to a reduction in hypothesis-only bias by comparing accuracy for each protocol for RoBERTa$_{\rm{Lg}}$ trained on hypothesis-only input, where lower accuracy suggests less bias in the data (Figure \ref{fig:indomain_val}). 
Both LitL and LitL Chat result in lower hypothesis-only bias than Baseline, and this gap widens in later rounds. 
To assess whether this widening from Round~1 to~5 is statistically reliable, we conduct a two-way ANOVA of round by protocol, which yields a significant interaction (\textit{p} = 0.049), indicating that while hypothesis-only performance increases for all protocols with more training examples, this increase in bias is significantly reduced in LitL and LitL Chat compared to Baseline.
The lower bias in LitL and LitL Chat may be due to the lower average word-label PMI, which increases over rounds for Baseline while consistently falling in both LitL and LitL Chat.\footnote{A two-way ANOVA again reveals a significant interaction of protocol by round (\textit{p} = 0.022) on word-label PMI values.} 
However, for all protocols, accuracies are still above chance performance, leaving room to further reduce hypothesis-only bias.

\paragraph{Diagnostic Sets}
We use diagnostic tests to evaluate whether fine-tuning on data collected with linguist involvement leads to a model that has higher performance on challenge test sets. 
Figure \ref{fig:iter_val} shows model performance on the GLUE diagnostic set and HANS non-entailment examples. 
A two-way ANOVA of round by protocol does not reveal any significant interactions or main effects for GLUE. 
For HANS, we see higher accuracy from LitL and LitL Chat for Lexical Overlap and Subsequence examples in Rounds 4 and 5 after introducing \textit{No} and \textit{All Overlap} constraints, though the interaction is only significant with RoBERTa$_{\rm{Lg+MNLI}}$ (\textit{p} = 0.0021 and \textit{p} = 0.0017 for Lexical Overlap and Subsequence, respectively). % despite the apparently larger accuracy increases in RoBERTa$_{\rm{Lg}}$. 
%This is likely due to greater variance in the data, indicating that there may be strong effects of individual annotators on lexical overlap and subsequence biases. 
Performance on HANS entailment examples are in line with \citet{mccoy-etal-2019-right} with median accuracies of 90\% or higher (Appendix \ref{app:hans_entailment}).

To investigate whether rates of lexical overlap differ by protocol, we assess classification accuracy for a linear model trained only on the example's overlap rate, defined as the proportion of words in the hypothesis that are also in the premise. We observe that the potential bias introduced from overlap rate is strongest in the Baseline protocol, which performs 9.52 points above majority class guessing, while LitL and LitL Chat perform 8.06 and 6.88 points above majority class guessing, respectively.

%While our interventions do not improve performance on more general examples from the GLUE diagnostic set, we do see reduced bias on measures related to overlap rates, and this is likely a result of direct targeting of that bias via the constraints of No Overlap and All Overlap.

\begin{figure}[h]
    \centering
    \includegraphics[width=0.5\textwidth]{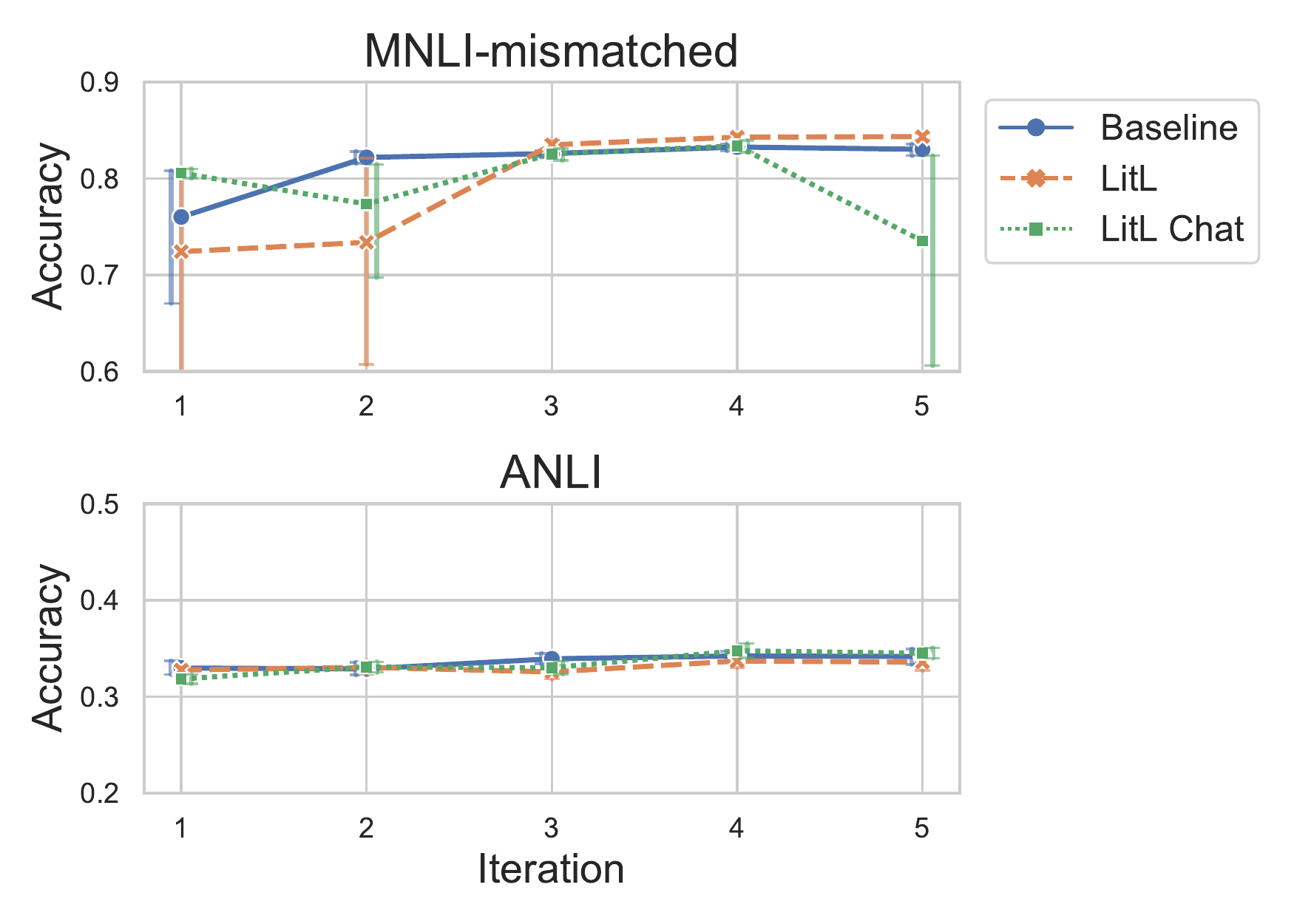}
    \caption{Performance of RoBERTa$_{\rm{Lg}}$ fine-tuned on data collected through different protocols on MNLI-mismatched (top) and ANLI (bottom).}
    \label{fig:combined_val}
\end{figure}

\paragraph{Held-Out Evaluation Sets}
We next test whether models fine-tuned on data collected with linguist involvement show better out-of-domain performance. % as a way of testing if linguistic input leads to data that allows for a greater degree of generalizability.
Figure \ref{fig:combined_val} shows that there is little difference in ANLI and MNLI-mismatched performance among all protocols. 
We perform a more granular analysis on ANLI examples using the tags from \citet{williams2020anlizing} and again find no clear effect of protocol (details in Appendix \ref{app:anli_by_tag}). 
Even though our interventions reduce hypothesis-only bias and improve model performance on HANS non-entailment examples, we have no evidence that these benefits transfer to out-of-domain examples or examples from adversarial protocols.

\section{Considerations in Choosing a Protocol} % Alicia

In broad terms, we observe a benefit from dynamically updating annotator guidelines to address gaps and biases observed during data collection.
This procedure increased the average cost per example by 4.1\% over an average base cost of \$0.367.
In an exit survey, %\footnote{Response rates on the exit survey were as follows -- Baseline: 83.8\%, LitL: 90.0\%, LitL Chat: 75.0\%}, 
over 50\% of annotators in LitL and LitL Chat indicated that they would have completed more optional challenge examples if the pay had been higher.
We offered \$0.05 to \$0.10 per example, but given the somewhat low rate at which annotators chose to attempt the challenges (28.6\% and 21.2\% of examples for LitL and LitL Chat, respectively), we find it likely that increasing the amount offered per example would have increased participation, potentially also increasing the benefits observed in model performance.
We recommend that future work using challenge options offer bonuses worth at least 15\% of the base pay. 

\paragraph{Cost of Linguist Involvement}
The iterative analyses and updates to the guidelines used for LitL and LitL Chat protocols took 10--12 hours of expert time per week, compared to a base time cost of about one hour per week for just monitoring task completion.
The use of Slack nearly doubled the expert time needed, adding an \textit{additional} 8--10 hours each week just for LitL Chat over LitL, even after taking into account the slight reduction in time spent replying to email questions that shifted to Slack. 
If we value linguist time at \$40/hr, then this raises the final price per example to \$0.378 in Baseline, with LitL 31.2\% higher, and LitL Chat 58.5\% higher than baseline. %\$0.496 in LitL, and \$0.599 in LitL Chat.
Despite the extra time spent with annotators, we did not observe any measurable benefit to this more hands-on intervention.

\paragraph{Qualitative Considerations} %of linguist involvement}
Though many annotators in LitL Chat expressed that they enjoyed the extra communication and help, annotators from LitL and LitL Chat rated the task as `more enjoyable' than typical MTurk tasks at nearly identical rates (85.2\% and 87.5\% respectively, compared to 67.7\% in Baseline).
Ratings of the difficulty of writing and validation tasks were also nearly identical between the LitL and LitL Chat protocols.

\section{Conclusion} % Alicia
We conclude that for some tasks, it is beneficial to integrate expert analysis of data \textit{during} data collection, so that the expert can dynamically update guidelines and constraints based on existing gaps and biases in the dataset.
Through a controlled experiment, we have shown that % dynamically updating targeted instructions \textit{does} improve training data. 
this expert involvement leads to higher accuracy in measures related to the dataset gaps and biases that were identified and targeted% during data collection
, and the iterative procedure allowed us to identify new areas of weakness at each round. 
As we did not observe any increases in out-of-domain accuracy, we conclude that %although our intervention results in a dataset that leads to higher in-domain accuracy, 
there is no more general benefit from our interventions beyond the targeted areas of weakness.
Future work could extend this protocol to identify additional interventions that would lead to datasets with better generalizability. % leaving open the question of how to design crowdworking protocols for better out-of-domain performance to future work.
%This procedure will therefore be most effective at improving data collection in cases where some sources of data bias are either already known or relatively straightforward to identify.
%We recommend incorporating high bonus rates for any optional constraints introduced, in order to increase the rate at which crowdworkers attempt to write more challenging annotations. 
%Finally, though some studies report that one-on-one interactions between experts and crowdworkers is beneficial in more challenging tasks, our direct comparison of protocols to test this claim no evidence to support it.
Finally, we find no evidence to support the claim reported by some studies that one-on-one interactions between experts and crowdworkers is beneficial in more challenging tasks.

%\section*{Acknowledgments}

\bibliographystyle{acl_natbib} % still correct for the emnlp 2021 template
\bibliography{anthology,LitL}

\newpage

%\phantom{...}

%\newpage

\appendix
\section{List of challenge options}
\label{challenge_options}
\paragraph{Lexical Options}
\begin{itemize}[]
    \item \textbf{Temporal reasoning} (Round 2): The hypothesis should reference two separate time points.% For example, the hypothesis may include words such as \textit{before} and \textit{after}, which shifts the time-point of the premise. 
    \item \textbf{Restricted word in different label} (Round 2): The hypothesis should contain a word that is banned for a different label.% For example, \textit{not}, which is restricted for contradiction sentences, is used for entailment sentences.
    \item \textbf{Hypernym or hyponym} (Rounds 2 \& 3): The hypothesis should contain a hypernym or hyponym (a more or less specific word or phrase) of a word in the premise.% For example, a more specific term for \textit{dog} is \textit{golden retriever}, and a less specific term is \textit{animal}.  
    \item \textbf{Synonym or antonym} (Rounds 2 \& 3): The hypothesis should contain a synonym or antonym of a word in the premise. %For example, a synonym of \textit{skinny} is \textit{thin} and an antonym is \textit{fat}.  
    \item \textbf{No overlap} (Rounds 4 \& 5): The hypothesis should use none of the content words appearing in the premise. Content words are nouns, verbs, adjectives, and adverbs. %This restriction includes not reusing proper nouns such as \textit{Frank Grimes} and \textit{Rome}. Instead, one could use common nouns such as \textit{the new employee} or \textit{a big city}.  
    \item \textbf{All overlap} (Rounds 4 \& 5): The hypothesis should only use content words that appear in the premise. Introducing new function words is allowed, as is changing grammatical features of the content words. %For example, given the premise \textit{The girl will climb the tree}, the hypothesis could be \textit{The tree will be climbed by the girl}.
\end{itemize}

\paragraph{Syntactic Options}
\begin{itemize}[]
    \item \textbf{Relative clause} (Round 2): The hypothesis should contain a relative clause. A relative clause is a noun that is described by a phrase that begins with words like \textit{who} or \textit{that}.
    \item \textbf{Reverse argument order} (Rounds 2 \& 3): The hypothesis should contain a pair of noun phrases from the premise in reverse order.% For example, \textit{Bill likes Jane} is the result of reversing the order of the nouns in \textit{Jane likes Bill}.
    \item \textbf{Grammar change} (Round 4): The hypothesis should change a grammatical element of the premise, such as tense, number, or gender on a pronoun.% by shifting the animacy or gender of the noun. %For example, given the premise \textit{Sally will leave the bar}, the hypothesis \textit{Sally has left the bar} is likely a contradiction.%Similarly, \textit{I saw her yesterday.} is true under different circumstances than \textit{I saw them yesterday.} In this case, the gender and number of the pronoun are changed.
\end{itemize}

\paragraph{World Knowledge Options}
\begin{itemize}[]
    \item \textbf{Background knowledge} (Rounds 2 \& 4): The hypothesis should target background facts or general knowledge that workers can infer from the premise. %For example, heaviness is a property of trucks in general, and a potato sack can hold things that are not potatoes. We use various titles for this challenge option such as `external knowledge' and `not obvious' to encourage workers to approach this constraint in different ways.  
    \item \textbf{Sub-part} (Round 3): The hypothesis should refer to something that is a part of an entity in the premise. For example, sub-parts of a \textit{bus} include its \textit{steering wheel}, and \textit{engine}.
    \item \textbf{Register change} (Round 5): The hypothesis should differ from the original text in its level of formality. %Register can either be elevated to `sound fancier' or lowered `as if the text were spoken to a childhood friend'. %This option was inspired by looking at the data and analyzing some of the creative examples that changed the register.  
\end{itemize}

\section{MNLI-Pretrained RoBERTa Results}
\label{app:mnli_trained_results}
\begin{figure}[h]
    \centering
    \includegraphics[width=0.5\textwidth]{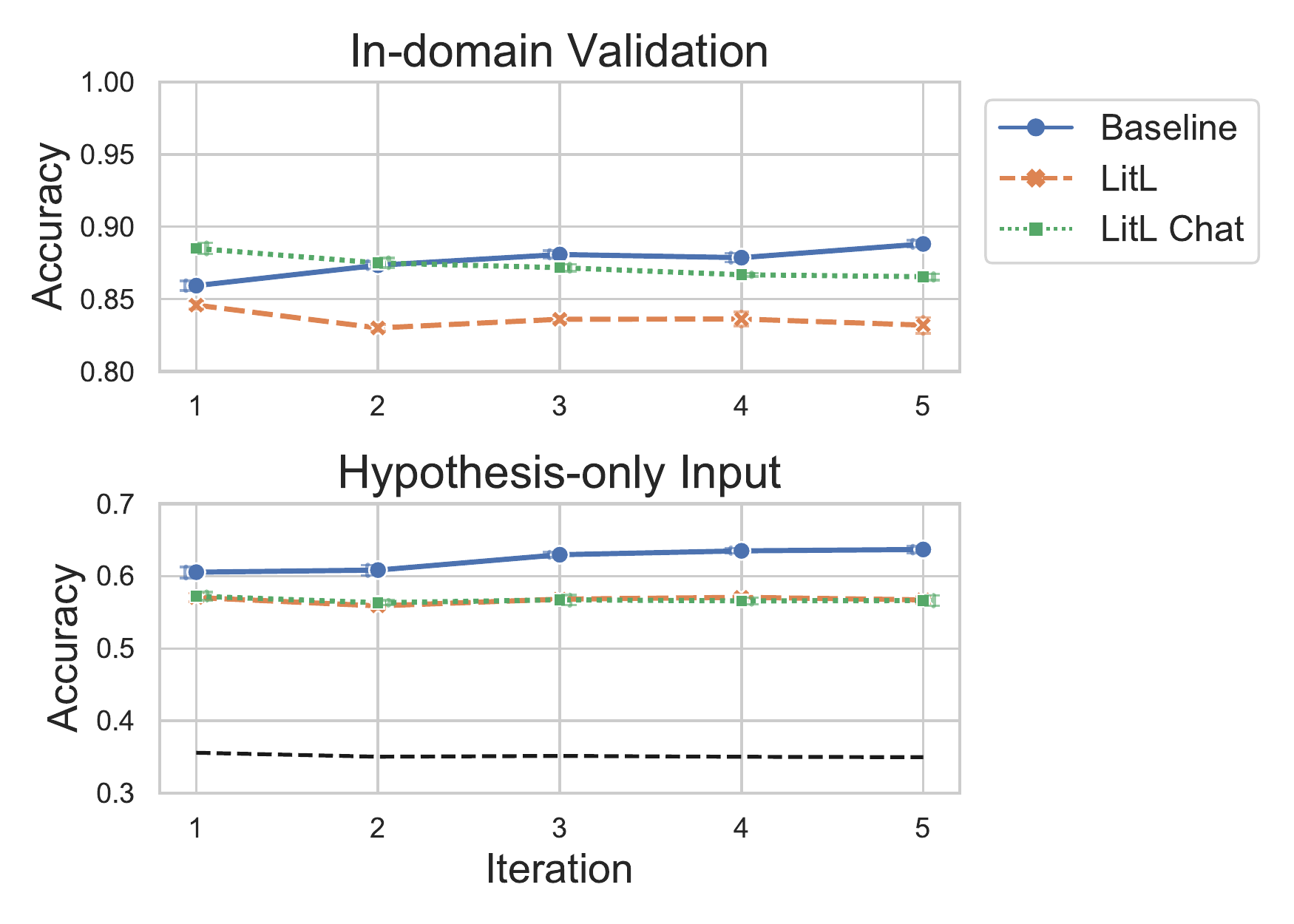}
    \caption{Performance of RoBERTa$_{\rm{Lg+MNLI}}$ fine-tuned on data collected through different protocols on in-domain validation data trained with either the full example (top) or hypothesis-only (bottom) input. Higher hypothesis-only accuracy indicates more bias. For each round, we include training and validation data \textit{accumulated} up to Round $n$. Dashed black line marks average majority class baseline across protocols.}
    \label{fig:roberta-mnli_indomain_val}
\end{figure}

\begin{figure}[h]
    \centering
    \includegraphics[width=0.5\textwidth]{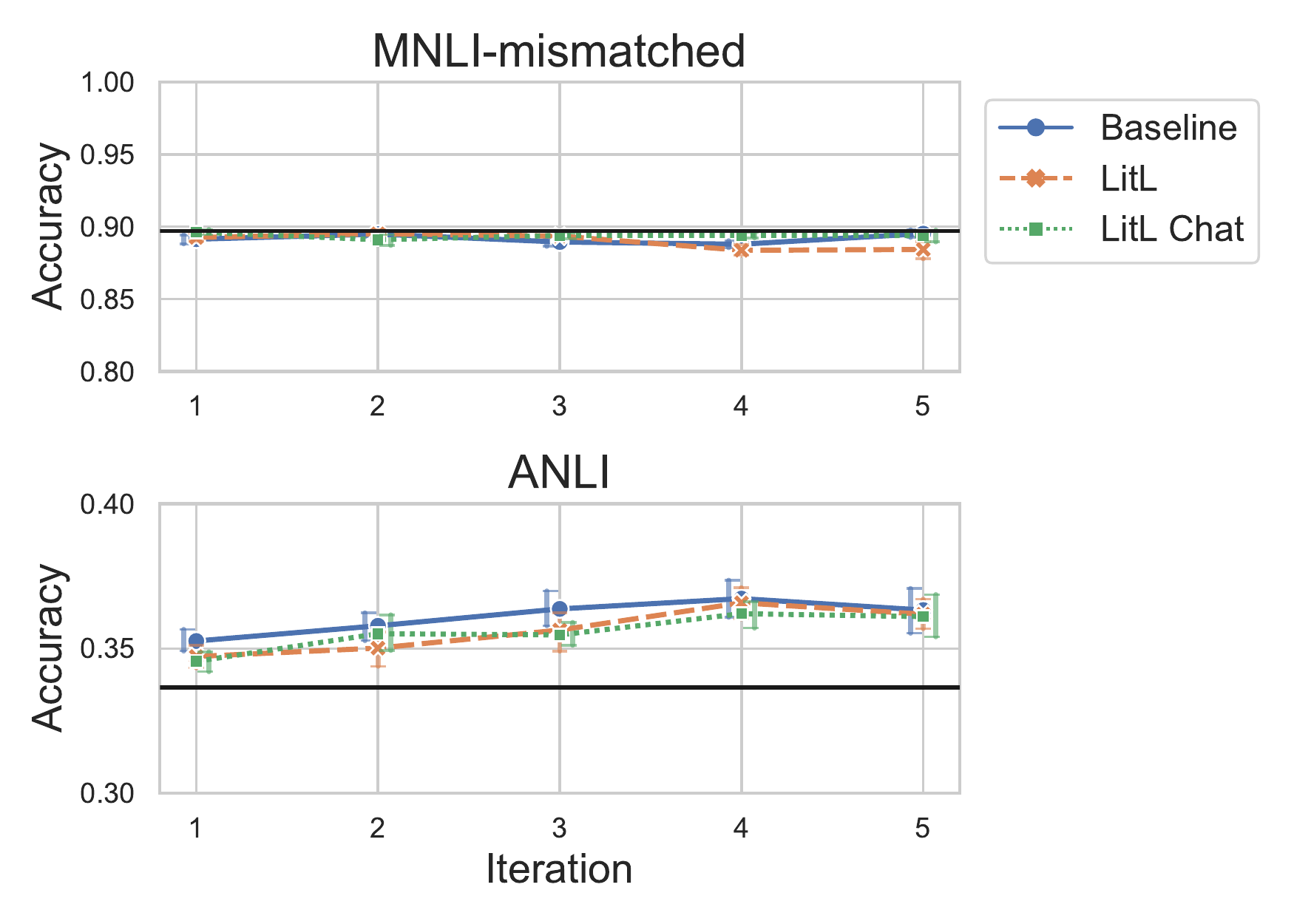}
    \caption{Performance of RoBERTa$_{\rm{Lg+MNLI}}$ fine-tuned on data collected through different protocols on MNLI-mismatched (top) and ANLI (bottom). The black line for MNLI-mismatched and ANLI indicates performance of RoBERTa$_{\rm{Lg}}$ fine-tuned on MNLI alone.}
    \label{fig:roberta-mnli_combined_val}
\end{figure}

\begin{figure*}[h]
    \centering
    \includegraphics[width=\textwidth]{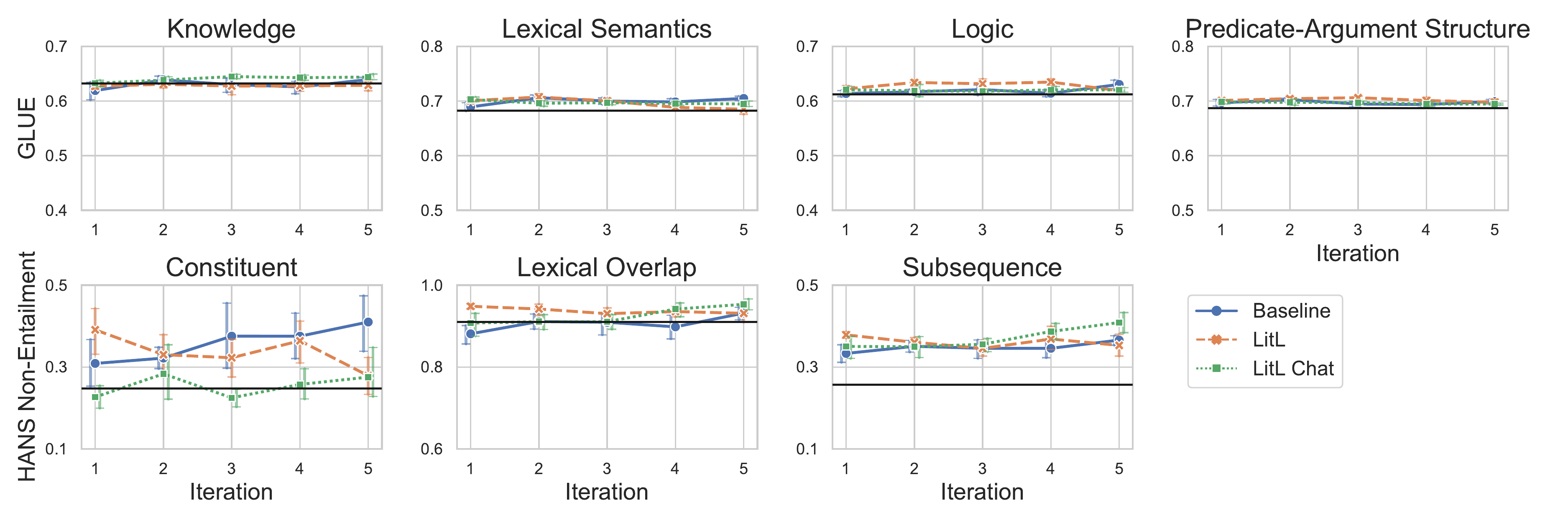}
    \caption{Performance of RoBERTa$_{\rm{Lg+MNLI}}$ fine-tuned on data collected through different protocols on the GLUE diagnostic set (top) and HANS non-entailment examples (bottom). The black line indicates performance of RoBERTa$_{\rm{Lg}}$ fine-tuned on MNLI alone.}
    \label{fig:roberta-mnli_iter_val}
\end{figure*}

\begin{figure*}[h]
    \centering
    \includegraphics[width=\textwidth]{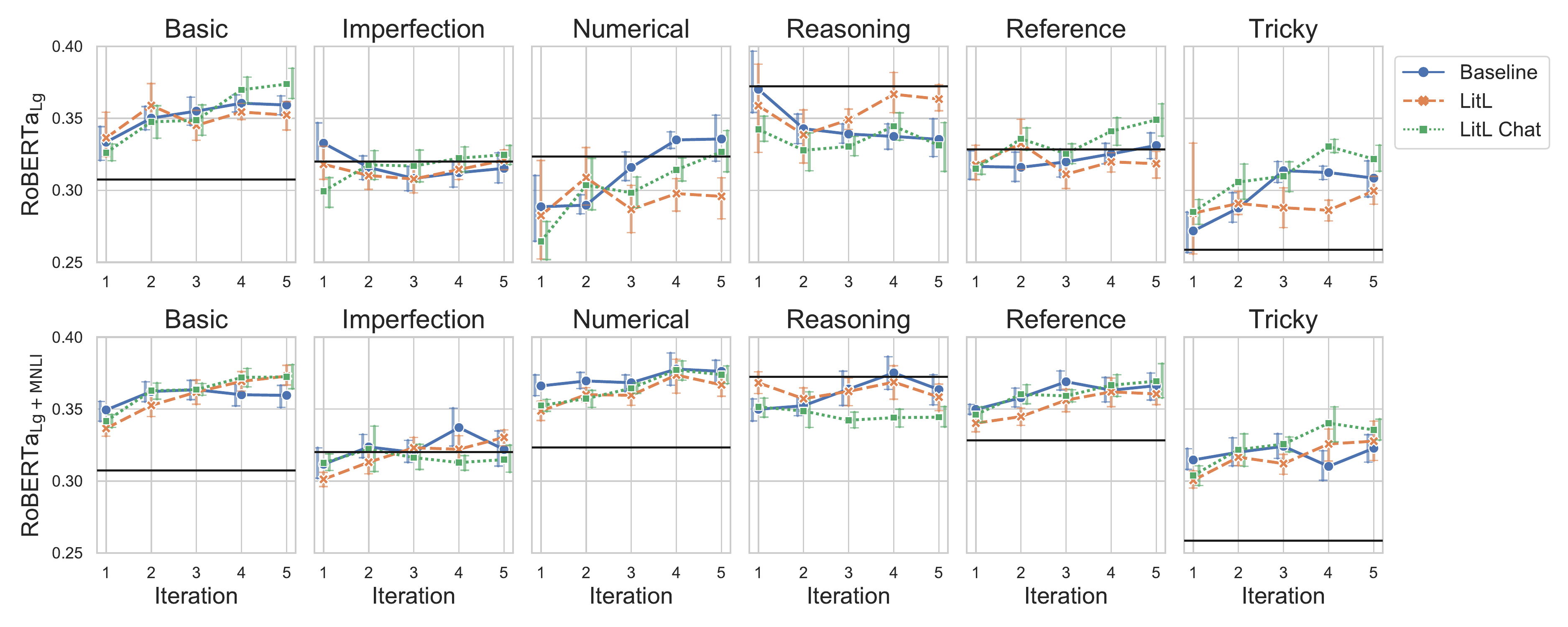}
    \caption{Performance of RoBERTa$_{\rm{Lg}}$ (top) and RoBERTa$_{\rm{Lg+MNLI}}$ (bottom) fine-tuned on data collected through different protocols on ANLI by reasoning tag from \citet{williams2020anlizing}. The black line indicates performance of a RoBERTa$_{\rm{Lg}}$ trained on MNLI \textbf{alone}.}
    \label{fig:anli_breakdown}
\end{figure*}

\begin{figure*}[h]
    \centering
    \includegraphics[width=\textwidth]{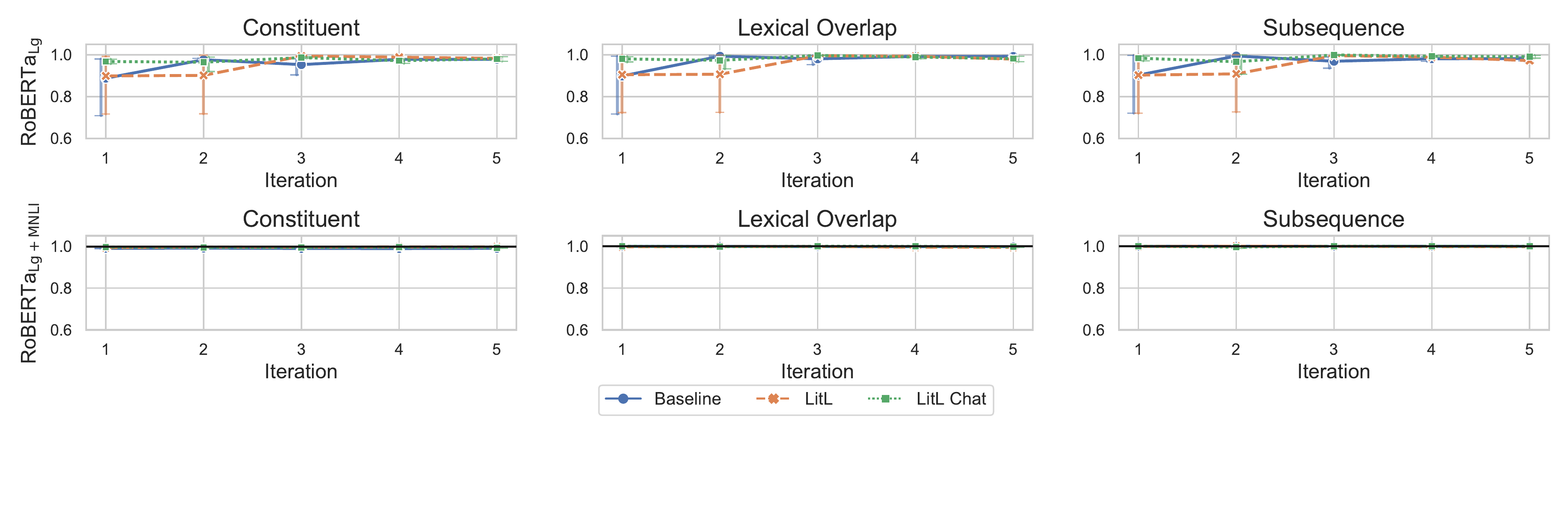}
    \caption{Performance of RoBERTa$_{\rm{Lg}}$ (top) and RoBERTa$_{\rm{Lg+MNLI}}$ (bottom) fine-tuned on data collected through different protocols on HANS entailment examples. The black line indicates performance of a RoBERTa$_{\rm{Lg}}$ trained on MNLI \textbf{alone}.}
    \label{fig:hans_entailment}
\end{figure*}

We fine-tune a RoBERTa$_{\rm{Lg}}$ model previously trained on MNLI (RoBERTa$_{\rm{Lg+MNLI}}$) on the same sets of training data used for the RoBERTa$_{\rm{Lg}}$ analyses. We find similar trends to those from fine-tuning RoBERTa$_{\rm{Lg}}$ and report them in the same set of plots here.

Figure \ref{fig:roberta-mnli_indomain_val} shows the performance of RoBERTa$_{\rm{Lg+MNLI}}$ fine-tuned using either the full example or hypothesis-only input. For both types of input, we see a performance gap between Baseline and our intervention protocols. We perform a two-way ANOVA of round by protocol to see if this performance gap significantly changes between rounds 1 and 5 and find a significant interaction (\textit{p} $<$ 0.001 for both full example and hypothesis-only input). For full example input, this indicates that our interventions create more challenging evaluation data. For hypothesis-only performance, Baseline performance increases while LitL and LitL Chat remain relatively unchanged, indicating that our interventions mitigate stronger hypothesis-only bias in NLI datasets as new data is collected.

Figure \ref{fig:roberta-mnli_combined_val} shows the performance of RoBERTa$_{\rm{Lg+MNLI}}$ fine-tuned on different protocols on MNLI-mismatched and ANLI. We find no significant difference among protocols for either held-out set.

Figure \ref{fig:roberta-mnli_iter_val} shows the performance of RoBERTa$_{\rm{Lg+MNLI}}$ fine-tuned on data from different protocols on the GLUE diagnostic set and HANS non-entailment examples. For the GLUE diagnostic set, we do not find any significant difference among protocols. For the HANS examples, we perform another two-way ANOVA of round by protocol and find significant interaction terms for all HANS categories (\textit{p} = 0.0018, 0.0021, 0.0017 for Constituent, Lexical Overlap, and Subsequence, respectively). For Lexical Overlap and Subsequence, these findings indicate our interventions lead to higher accuracy compared to Baseline. However, we see the opposite from Constituent examples with both intervention protocols performing worse than Baseline.

\section{ANLI Performance by Reasoning Type}
\label{app:anli_by_tag}

We test whether \textit{any} of the reasoning tags in ANLI \citep{williams2020anlizing} reveal an area where data collection with linguist involvement leads to improved model performance.
Figure \ref{fig:anli_breakdown} shows the performances of RoBERTa$_{\rm{Lg}}$ and RoBERTa$_{\rm{Lg+MNLI}}$ on ANLI by reasoning tag. 
Similar to our findings in Figures \ref{fig:combined_val} and \ref{fig:roberta-mnli_combined_val}, we do not find any increases in accuracy from our interventions for any reasoning tags.

\section{HANS Entailment Peformance}
\label{app:hans_entailment}

On the entailment subset of HANS, models typically achieve accuracies near 100\% \citet{mccoy-etal-2019-right}. 
This is because the three heuristics in HANS target instances that lead to a greater likelihood of the model choosing \textsc{entailment} compared to \textsc{neutral} or \textsc{contradiction}, and thus the non-entailment portion of HANS is the challenge set.
Figure \ref{fig:hans_entailment} shows the performance of RoBERTa$_{\rm{Lg}}$ and RoBERTa$_{\rm{Lg+MNLI}}$ fine-tuned on our data and tested on HANS entailment examples. 
For RoBERTa$_{\rm{Lg}}$, variability in performance reduces in later rounds as the training set size grows with 3k examples per round, though median performances for all rounds are still 90\% or higher. 
For RoBERTa$_{\rm{Lg+MNLI}}$, accuracies are near 100\%, consistent with \citeauthor{mccoy-etal-2019-right}'s findings.

\section{Examples of Collected Data}
\label{app:examples}

In order to show a representative sample of the validated data, we randomly sample premises from Round 5 data for which annotations exist in all three labels for each protocols (roughly 45\% of that round's validated data).
Five such examples are presented in Table \ref{tab:examples}.
Example complexity varies widely from example to example, and it is not always the case that the example in Baseline is the simplest one.
For premise 4, for example, the Baseline crowdworker has written very complex examples that require abstract reasoning about the knowledge that \textit{Harris} has.
For this same premise, the LitL Chat crowdworker has also created a tricky set of examples, in this case ones that do not re-use any words from the original premise.

In premise 3, we see an example where the LitL Chat crowdworker uses the idiom \textit{seen better days} for the entailment example, in place of just using a different lexical item for \textit{tough} as the crowdworkers in the other two protocols do.
Use of idioms was suggested to workers in LitL and LitL Chat as one way to write more creative examples.
In premise 5, we see that the LitL crowdworker has written a challenging contradiction example, one which requires knowledge that if help is needed on a project, that means it must not be complete.

\begin{table*}[t!]
    \newcolumntype{z}{>{\hsize=.03\hsize}X}
    \newcolumntype{s}{>{\hsize=.15\hsize}X}
    \newcolumntype{m}{>{\hsize=.75\hsize}X}
    \centering
    \footnotesize
    \resizebox{\textwidth}{!}{% 
    \begin{tabularx}{\textwidth}{zmsXXX}
    %{p{0.12\textwidth}|c|p{0.25\textwidth}p{0.25\textwidth}p{0.25\textwidth}}
        \toprule
        & Premise & Label & \multicolumn{3}{c}{Hypothesis}  \\
        & & & \multicolumn{1}{c}{Baseline} & \multicolumn{1}{c}{LitL} & \multicolumn{1}{c}{LitL Chat} \\
        \midrule
        1 & \multirow{3}{\hsize}{(The Ramseys buried their daughter in Atlanta, then vacationed in Sea Island, Ga.) This absence, some speculate, gave the Ramseys time to work out a story to explain their innocence.} & \cellcolor{gray!25}E & \cellcolor{gray!25} Some people were skeptical of the Ramseys' reasons for going on vacation. & \cellcolor{gray!25}The Ramseys came up with a story to tell the media they didn't do it. & \cellcolor{gray!25}Some speculate that the Ramseys worked out a story while on vacation. \\
        & & N & The Ramsey's held a private funeral service for their daughter. & The Ramseys had nothing to hide. & The Ramseys worked in Atlanta. \\
        & & \cellcolor{gray!25}C & \cellcolor{gray!25}The Ramsey's daughter joined them on their trip to Sea Island. & \cellcolor{gray!25}The Ramseys went into mourning after burying their daughter. & \cellcolor{gray!25}The Ramseys buried their daughter in Sea Island, Ga.	 \\
        & &\cellcolor{gray!25}&\cellcolor{gray!25}&\cellcolor{gray!25}&\cellcolor{gray!25}\\
        &&&&&\\
         \midrule
        2 & \multirow{3}{\hsize}{Mr. Clinton rewards Mr. Knight for his fund raising, Mr. Gore lays the groundwork for his anticipated presidential bid four years from now, and the companies, by hiring Mr. Knight, get the administration's ear.} & E & Al Gore planned to run for president. & Mr. Gore lays the groundwork for his anticipated presidential bid four years from now. & By hiring Mr. Knight, companies were listened to by the administration.	\\
        & & \cellcolor{gray!25}N & \cellcolor{gray!25}Companies were hopeful they could get Clinton to further reduce corporate tax rates. & \cellcolor{gray!25}Mr. Knight get the administration's ear for companies that contribute to his fund raising. & \cellcolor{gray!25}The administration had been ignoring the companies up to this point. \\
        & & C & Bill Clinton punished Mr. Knight because of his fund raising efforts. & Mr. Clinton admonishes Mr. Knight for his fund raising. & Companies were ignored by the adminstration because of the hiring of Mr. Knight. \\
         \midrule
        3 & \multirow{3}{\hsize}{And these are tough times for reviewers in general.} & \cellcolor{gray!25}E & \cellcolor{gray!25}Reviewers are going through difficult times. & \cellcolor{gray!25}Reviewers are having a challenging time. & \cellcolor{gray!25}Reviewers have seen better days. \\
        & & N & The recession is to blame for these tough times. & Times will only get tougher for reviewers. & Reviewers are still able to get by. \\
        & & \cellcolor{gray!25}C & \cellcolor{gray!25}This is a great time to be a reviewer. & \cellcolor{gray!25}Reviewers have rarely had it so easy. & \cellcolor{gray!25}This have to be the best time to get into the review game. \\
         \midrule
        4 & \multirow{3}{\hsize}{To some critics, the mystery isn't, as Harris suggests, how women throughout history have exploited their sexual power over men, but how pimps like him have come away with the profit.} & E & The author argues that some critics are incapable of understanding the role pimps have played in the exploitation of women.	 & pimps like him have profited.  & An unsolved question involves the money making of a hustler.	 \\ 
        & & \cellcolor{gray!25}N & \cellcolor{gray!25}If women are going to attempt to exploit their sexual power over men, then it is only natural for pimps to emerge to oversee sexual transactions. & \cellcolor{gray!25}Pimps have exploited women who have more power than they think. & \cellcolor{gray!25}Reviewers are mainly concerned with hustlers.	\\ 
        & & C & Harris does not understand the means by which women have using sexual power in order to exploit men. & Pimps control every woman. & An unsolved question involves the money wasting of a hustler. \\ 
         \midrule
        5 & \multirow{3}{\hsize}{We need your help with another new feature that starts next week.} & \cellcolor{gray!25}E & \cellcolor{gray!25}Next week, a new feature will be introduced. & \cellcolor{gray!25}There have been other new features. & \cellcolor{gray!25}We are starting a new feature next week. \\ 
        & & N & This new feature focuses on cloud technology. & Help has been needed with previous features. & We are starting a new feature next week that uses maps. \\ 
        & & \cellcolor{gray!25}C & \cellcolor{gray!25}The new feature will start six months from now. & \cellcolor{gray!25}The project is complete and currently unsupported. & \cellcolor{gray!25}We have more help than we need for the new feature next week. \\ 
         \bottomrule
    \end{tabularx}
    }
    \caption{Randomly selected examples from validation data showing typical writing from each protocol.}
    \label{tab:examples}
\end{table*}

\end{document}